\documentclass[twocolumn, switch]{article} 

\usepackage{preprint}

\usepackage{amsmath, amsthm, amssymb, amsfonts}

\usepackage[numbers,square]{natbib}
\usepackage{hyperref}	

\DeclareMathOperator{\avg}{avg}
\DeclareMathOperator{\Softmax}{Softmax}

\usepackage{titlesec}
\titlespacing\section{0pt}{12pt plus 3pt minus 3pt}{1pt plus 1pt minus 1pt}
\titlespacing\subsection{0pt}{10pt plus 3pt minus 3pt}{1pt plus 1pt minus 1pt}
\titlespacing\subsubsection{0pt}{8pt plus 3pt minus 3pt}{1pt plus 1pt minus 1pt}

\title{MuSeM: Detecting Incongruent News Headlines using Mutual Attentive Semantic Matching}

\begin{document}

	

	\author{Rahul Mishra \\ \textit{University of Stavanger, Norway} \\ {\texttt\small rahul.mishra@uis.no} \\ \And
		Piyush Yadav \\ \textit{Lero, NUI Galway, Ireland} \\ {\texttt\small p.yadav1@nuigalway.ie}\\ \And
		Remi Calizzano \\ \textit{DFKI, Germany}\\
		{\texttt\small remi.calizzano@dfki.de}  \And
		Markus Leippold \\ \textit{University of Zurich, Switzerland} \\
		{\texttt\small markus.leippold@bf.uzh.ch} \\ 
	}
	
	\date{}

	\twocolumn[ 
	\begin{@twocolumnfalse} 
		
		\maketitle
		
		\begin{abstract}
			Measuring the congruence between two texts has several useful applications, such as detecting the prevalent deceptive and misleading news headlines on the web. Many works have proposed machine learning based solutions such as text similarity between the headline and body text to detect the incongruence. Text similarity based methods fail to perform well due to different inherent challenges such as relative length mismatch between the news headline and its body content and non-overlapping vocabulary. On the other hand, more recent works that use headline guided attention to learn a headline derived contextual representation of the news body also result in convoluting overall representation due to the news body's lengthiness. This paper proposes a method that uses inter-mutual attention-based semantic matching between the original and synthetically generated headlines, which utilizes the difference between all pairs of word embeddings of words involved. The paper also investigates two more variations of our method, which use concatenation and dot-products of word embeddings of the words of original and synthetic headlines. We observe that the proposed method outperforms prior arts significantly for two publicly available datasets. 
		\end{abstract}
		\vspace{0.35cm}
		
	\end{@twocolumnfalse} 
	] 

	\section{Introduction}
	In the age of the prevalence of smart handheld devices, most of the information consumption is digital. This paradigm shift in the way people consume information has also brought forth several new challenges, such as misinformation and deceptive content. News headlines that incorrectly represent the content of the news body are called incongruent or click-baits. A deceptive and misleading news headline can result in false beliefs and wrong opinions. News titles play an essential role in making first impressions to readers and thereby deciding the viral potential of news stories within social networks \cite{Reis}. Most users rely only on the news title content to determine which news items are significant enough to read \cite{Gabielkov}. The curse of deceptive content gets amplified by several magnitudes when people share it without reading news body content. 
	\begin{figure}[htbp]
		\begin{minipage}{0.5\textwidth}
			\centering
			\centering
			\vspace*{-12pt}
			\hspace*{-0mm}\includegraphics[scale=0.13]{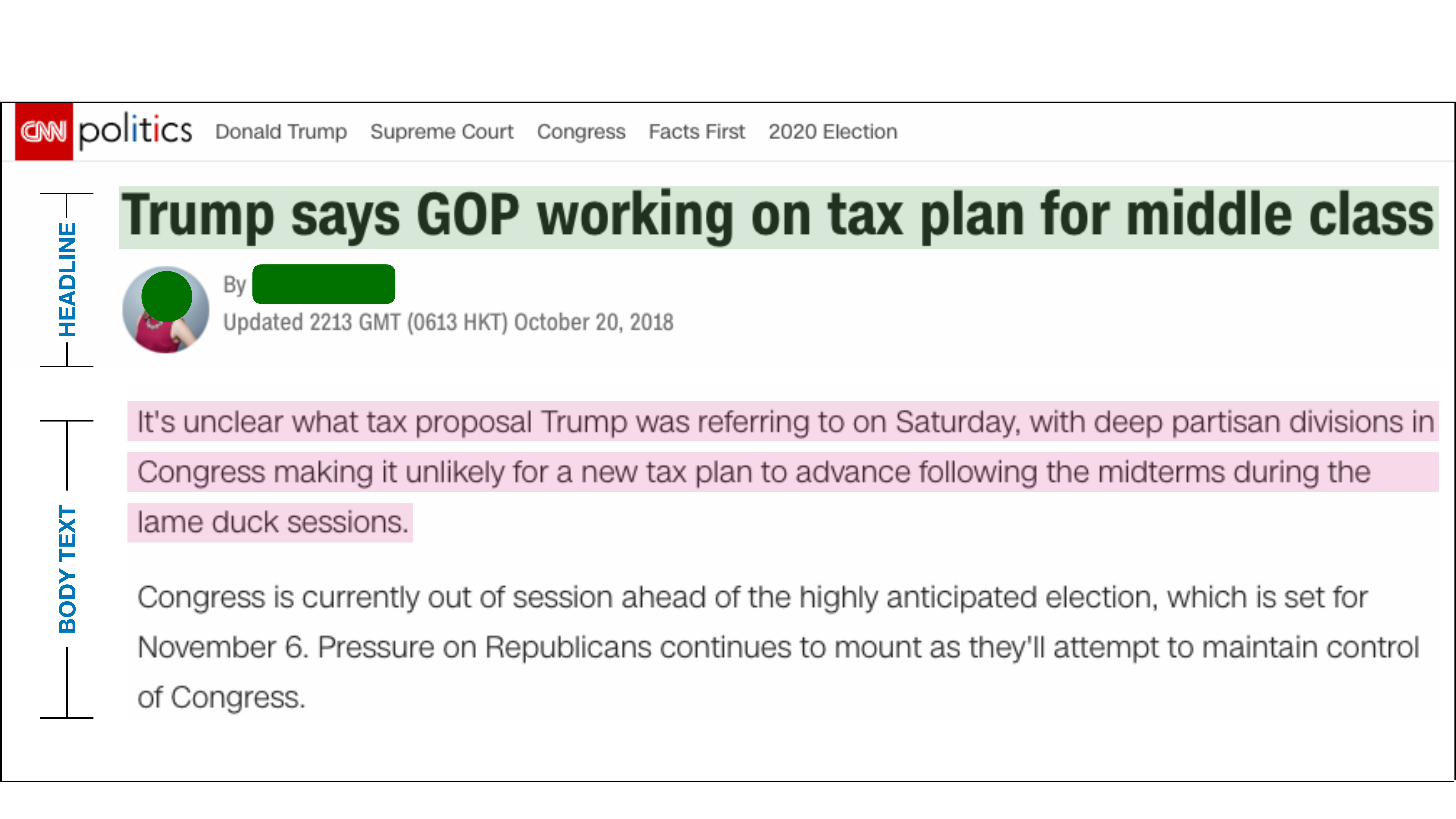}
			
			
			\label{fig:example}
		\end{minipage}
		\begin{minipage}{0.5\textwidth}
			\centering
			\hspace*{-0mm}\includegraphics[scale=0.132]{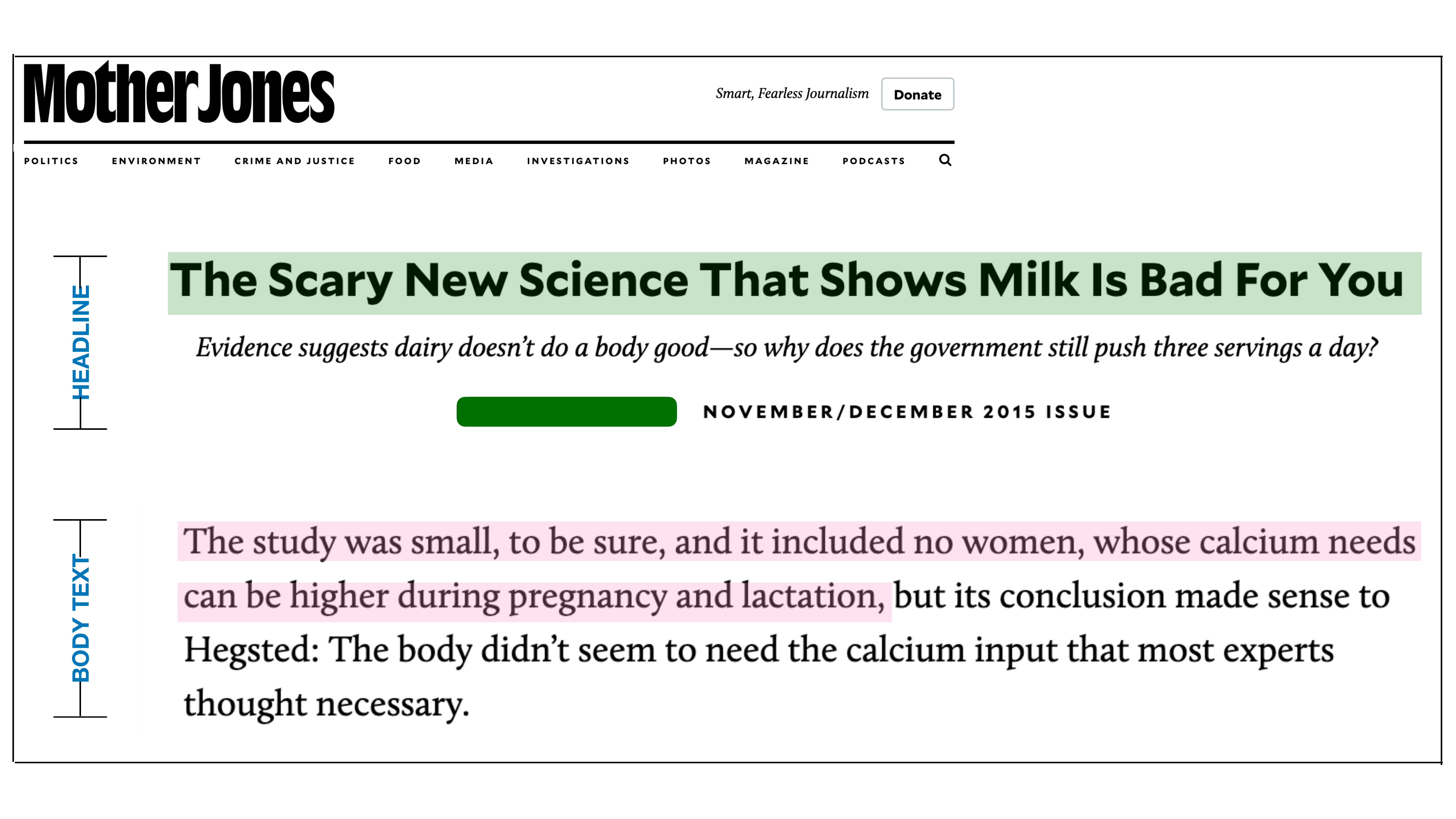}
			\caption{Examples of Incongruent Headlines related to politics and health-care.}
			
			\label{fig:example}
		\end{minipage}   		 
	\end{figure}
	Consider an example of an incongruent headline in Figure \ref{fig:example} taken from CNN (cnn.com).\footnote{\url{https://edition.cnn.com/2018/10/20/politics/donald-trump-tax-middle-income/index.html} } The headline states that ``Trump says GOP working on tax plan for middle class" whereas body content mentions that ``It's unclear what tax proposal Trump was referring to on Saturday" which contradicts with the headline.
	
	In another example in Figure \ref{fig:example},\footnote{\url{https://www.motherjones.com/environment/2015/11/dairy-industry-milk-federal-dietary-guidelines/}} the headline reads ``The Scary New Science That Shows Milk Is Bad For You" but a part of the body text clearly states that ``The study was small, to be sure, and it included no women." The headline radically generalizes the claim made in the study and exaggerates it.
	
	Many machine learning based solutions have been investigated previously in the literature to detect click-baits and news headline incongruence. Some initial works \cite{ferreira-vlachos-2016-emergent}\cite{ijcai2017-579} use sentence matching based methods to compute similarity or dissimilarity between the web claims and news headlines to detect incongruence. 
	Researchers \cite{Blom}\cite{Rony2017} have also utilized different classification methods and used linguistic and stylistic features to learn a classifier to identify the incongruity between news titles and news body. The authors in  \cite{Yoon} propose neural attention-based methods to find the entailment between the news items' title and body.  Some recent works \cite{8594871} have identified the significance of generative methods such as generative adversarial networks for incongruence detection.

	Previous methods applied to click-bait detection that use natural language processing techniques are not suitable for the detection of headline incongruence as this problem requires more facets and aspects to be covered than just stylistic features \cite{chesney-etal-2017-incongruent}. The methods that use a text similarity based approach to detect incongruence perform poorly because of the long text content of the news body. Text similarity schemes work well in case of short texts.

	This paper proposes a semantic matching technique based on inter-mutual attention that uses a synthetically generated headline corresponding to the news body content and original news headline to detect the incongruence.  
	The proposed inter-mutual attention technique is inspired by some recent works, which use the intra-mutual word to word attention within a sentence to detect sarcasm \cite{Tay2018} and intra-mutual user to user attention within a retweet propagation path sequence to detect misinformation \cite{Mishra_2020_CVPR_Workshops}.
	
	\begin{figure*}[t!!!]
		\centering
		\hspace*{-0mm}\includegraphics[scale=0.45]{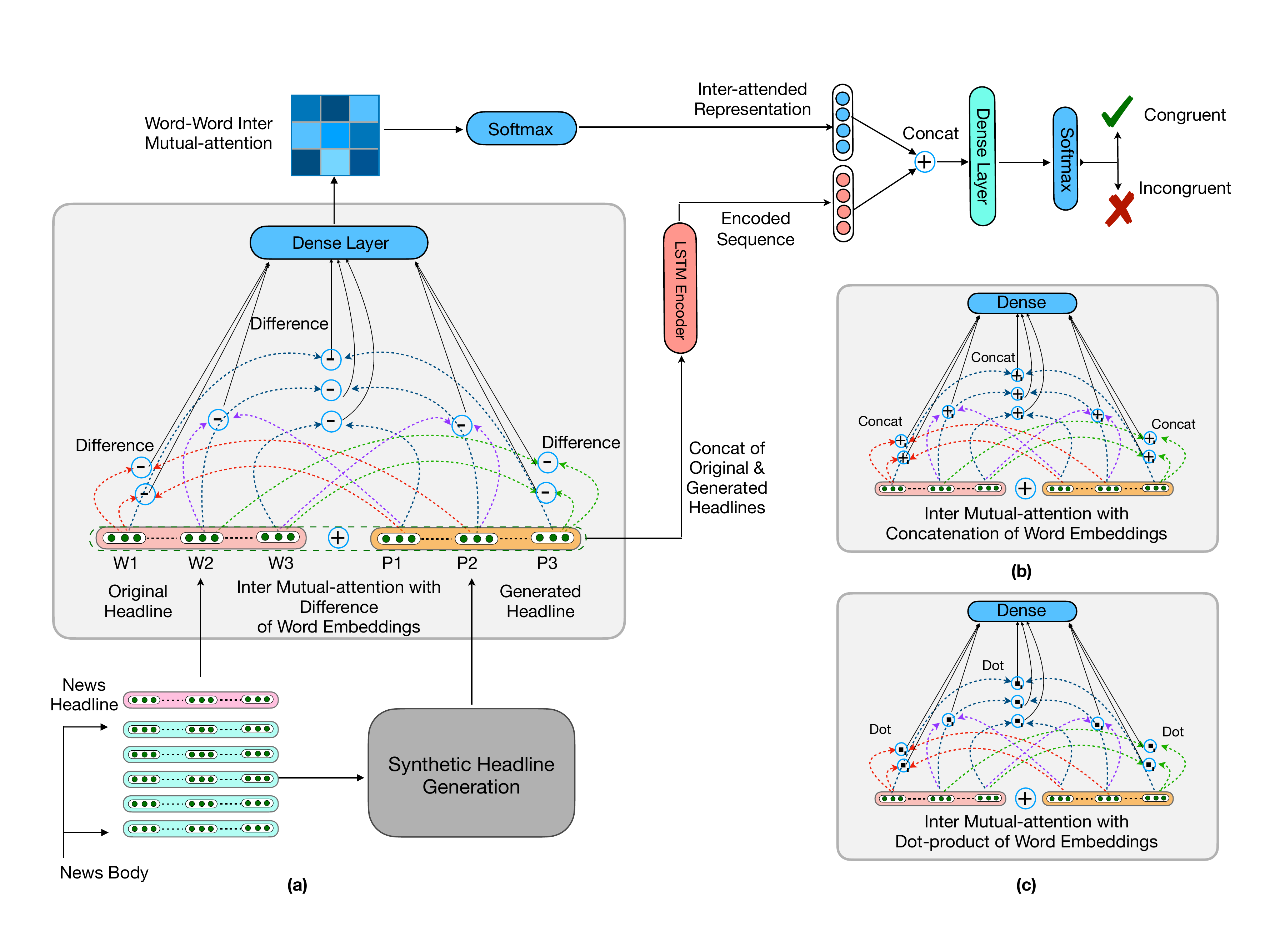}
		\caption{Overall Architecture of the MuSeM model. (a) A version of the MuSeM model with difference operation between word embedding pairs. (b) A depiction of concatenation operation between word embedding pairs. (c) A depiction of the dot-product operation between word embedding pairs. }
		
		\label{fig:attention}
		
	\end{figure*}

	In sum, the major contributions of this work are as follows:
	\begin{enumerate}
		\item We are the first to use inter-mutual attention based semantic matching to detect incongruent news headlines. The key idea is to get the pairwise difference between word embeddings of original and synthetic headlines and compute a mutual attention matrix after applying a dense layer. Subsequently, row-wise max-pooling can be used to compute the attention scores, to be used for the classification.       
		\item We use synthetic headlines generated from various generative adversarial networks based schemes using news body content to get the effective, contextual, and low dimensional representation.
		\item We also investigate two additional variants of the proposed model, which incorporate addition and concatenation of word embeddings of the word pairs of original and synthetic headlines.
		\item We combine all the three variants of word embedding operations in a clubbed model, which outperforms the three variants individually.
		\item We conduct experiments with two publicly available datasets, which show the effectiveness of the proposed models to detect incongruent news headlines.
		
	\end{enumerate}

	\section{Related Works} 
	Most of the literature's initial works propose using linguistic and statistical features based classifiers to detect click-baits and incongruent headlines. The authors of \cite{Yimin} suggest to perform lexical and syntactic analysis for the identification of the click-baits. In \cite{7752207}, authors use linguistic features to learn support vector machine (SVM) based classifier to detect click-baits, and they also release a manually annotated dataset. Authors of \cite{Potthast} use text features and meta-information of tweets to learn a classifier to detect click-baits. 
	
	On the other hand, some works \cite{ferreira-vlachos-2016-emergent}\cite{ijcai2017-579} deal with sentence matching based stance classification which is a closely related problem to headline incongruence. These methods are not suitable directly for headline incongruence due to some challenges such as relative length mismatch between the news headline and its body content and non-overlapping vocabulary. Authors in \cite{Wei} propose to use a co-training approach with myriad kinds of features such as, e.g., sentiments, textual, and informality. Some very recent works such as \cite{Yoon} use neural attention \cite{Bahdanau2015} based approach to achieve headline guided contextual representation of the news bod text.  This headline guided attention also results in convoluting the body content's overall representation due to its lengthiness.
	
	We propose an improved semantic matching between the news headline and its body text via inter-mutual attention. The inter-mutual attention based matching is performed between the original and a synthetically generated headline, rather than between original news headline and its body content. The synthetic headline is generated via state-of-the-art generative adversarial methods using the body contents of the news item. The generative adversarial techniques have been very successful in generating realistic artificial images \cite{Karras}, videos \cite{Suwajanakorn}\cite{Nagano} and text \cite{Jiaxian} contents. One of the pioneer work in artificial text generation is \cite{Lantao}, which solves the problem of generator differentiation by updating the gradient policy directly. The authors of \cite{Jiaxian} introduce a GAN based text generation method, which uses a third module, called MANAGER, which helps the generator network to utilize some leaked feature information from the discriminator network. In \cite{8594871}, authors introduce a synthetic headline generation method based on style transfer and generative adversarial network. 
	
	The key idea behind the inter-mutual attention method is inspired by two recent contributions  \cite{Tay2018}\cite{Mishra_2020_CVPR_Workshops}. There are some significant differences between these works and the proposed model. Firstly, they use intra-mutual word to word or user to user attention, which builds on word embedding pairs or user embedding pairs occurring within a sentence or retweet path sequence correspondingly to compute the attention scores. In contrast, the proposed model uses pairs of word embeddings of words that belong to two different sentences, i.e., original and synthetic headline. Secondly, these previous methods use a concatenation of embedding vectors to form an overall pair embedding representation. In contrast, the proposed method uses the difference between the word embeddings to form an overall pair embedding representation. The intuition behind these differences is evident as we detect incongruence between two pieces of texts (headline and body), and computing the difference between word embedding pairs results in capturing the cues related to semantic similarity or dissimilarity between the two.

	\section{Problem Definition and Proposed Model}
	This section introduces the problem definition and presents the overall architecture of the proposed model MuSeM in detail. We describe the synthetic headline generation schemes, and then we provide details of the inter-attentive semantic matching technique. We also present two more variants of MuSeM model and devise a clubbed model, which combines all the three variants.

	\subsection{Problem Definition}
	Given a news item $ n_i\in N$, having headline $h_i$ and body content $b_i$, we need to predict whether there is incongruence between $h_i$ and $b_i$ by classifying the news as either ``Congruent(C)'' or ``Incongreunt(I).'' The news headline $h_i$ consists of a sequence of $l$ words denoted as $h_i = \{w_{h1}, w_{h2}, ..., w_{hl}\} \in W$, and the news body content $b_i$ consists of a sequence of $m$ words denoted as $b_i = \{w_{b1}, w_{b2}, ..., w_{hm}\} \in W$.

	\subsection{Word Embedding Layer}
	For a news item $n_i$, the corresponding headline $h_i$ of length $l$ and body content $b_i$ of length $m$ are represented as $h_i = \{f(w_{h1}), f(w_{h2}), ..., f(w_{hl})\}$ where $\forall j, f(w_{hj}) \in \mathbb{R}^d$ is a word embedding vector of dimension $d$ for the $j^{th}$ word in headline $h_i$, and $b_i = \{f(w_{b1}), f(w_{b2}), ..., f(w_{hm})\}$ where $\forall k, f(w_{bk}) \in \mathbb{R}^d$ is a word embedding vector of dimension $d$ for the $k^{th}$ word in body content $b_i$. We experiment with pre-trained \textsc{glove} embeddings for the evaluation. 
	
	\subsection{Synthetic Headline Generation}
	As discussed in the introduction, the semantic similarity between the headline and news body content is the most significant incongruence indicator. Text similarity schemes work well in case of short texts. However, the major bottleneck to compute such a similarity measure is the news body's text length. For this reason, the methods which use a text similarity based approach to detect incongruence perform poorly. Generative adversarial networks (GANs) are recently getting much traction for various application scenarios, from the generation of realistic artificial images to compositions of meaningful poems. We utilize GANs based text generation techniques to generate a short synthetic headline for each of the news items using respective news body contents. The resulting synthetic headline is a low dimension contextual representation of the news body content.   
	
	\subsubsection{Generative Adversarial Network}
	A typical generative adversarial network \cite{Goodfellow} comprises two neural nets, a discriminator and a generator. Both of these sub neural nets compete with each other. The generator network tries to maximize the classification error, while the discriminator network tries to minimize it. Both the generator and discriminator converge together and reach an equilibrium state.

	\subsubsection{Sequence Generative
		Adversarial Nets (SeqGAN)}
	SeqGAN\cite{Lantao} is a sequence generation method based on generative adversarial networks. The key difference between standard GANs and SeqGAN is that SeqGAN bypasses the generator differentiation problem associated with original GANs by directly performing a gradient policy update using a reinforcement learning (RL) based approach. In the $\theta$-parameterized generative model $G_\theta$ and start state $s_0$, the expected end reward for output $y_1$ is defined as:
	\begin{equation}
	\label{eq:5}
	\begin{aligned}
	J(\theta) = E[R_T |s_0, \theta] = \sum_{y_1\in Y} G_\theta (y_1|s_0) \cdot Q^{G_\theta}_{D_\phi}(s_0, y_1),
	\end{aligned} 
	\end{equation}
	where $Q^{G_\theta}_{D_\phi}(s_0, y_1)$ is an action-value function. A recurrent neural network (RNN) is used for the generative model for sequences, and a convolutional neural network (CNN) is used as the discriminator. We refer to the respective paper for more details.

	\subsubsection{Stylized Headline Generation (SHG)}
	Since incongruent headlines and click-baits usually follow a certain catchy and deceptive writing style, it can be advantageous to generate synthetic headlines that mimic these similar writing styles. Authors of very recent work SHG\cite{8594871} propose a style transfer based headline generation approach, which uses a generative adversarial network with style discriminator. A gated recurrent unit (GRU) based recurrent neural network (RNN) is used as a generator that tries to minimize the following negative log-likelihood:
	\begin{equation}
	\label{eq:5}
	\begin{aligned}
	L_G(\theta_G) = E_{(x,h)\in S}[-\log_{PG}(h|y^L, z)].
	\end{aligned} 
	\end{equation}
	Three variants of discriminators are used, one for distinguishing the styles of the original and generated headline, a second one to maintain the aligned distributions in both the original and generated headlines, and a third for making sure that headline and body pairs are correctly classified. Again, we refer to the respective paper for more details.
	\subsection{Inter-mutual Attentive Semantic Matching}
	We now discuss the proposed semantic sentence matching scheme in detail. Our key idea is to model the relationships between all possible pairs of the words occurring in both the sentences, which captures the inherent semantic similarity between the sentences. We use pretrained word embeddings vectors and apply a novel inter mutual attention technique to model the similarity or dissimilarity relationship between pair of sentences. 
	\subsubsection{Word to Word Inter Mutual Attention}
	We compute mutual attention scores between a pair of sentences, of which first $h^o_i = \{f(w_{h^o1}), f(w_{h^o2}), ..., f(w_{h^ol})\}$ where $\forall j, f(w_{h^oj}) \in \mathbb{R}^d$  is the original headline of the news item, represented in terms of a sequence of word embeddings, and second $h^s_i = \{f(w_{h^s1}), f(w_{h^s2}), ..., f(w_{h^sp})\}$ where $\forall j, f(w_{h^sj})\in\mathbb{R}^d$  is the synthetically generated headline for news item $n_i$, represented in terms of a sequence of word embeddings. Here, $l$ and $p$ are lengths of the original headline $h^o_i$  and the synthetic headline $h^s_i$, respectively. 
	
	First of all, we compute the difference between word embedding vectors of each candidate word pair $W_q,W_r$. Candidate pairs are formed by selecting all possible combinations of the inter sentence word pairs, such as $W_q,W_r$, where $W_q$ and $W_r$ are the $q^{th}$ word of the original headline $h^o_i$ and the $r^{th}$ word of the synthetic headline $h^s_i$, respectively.  Now we use a dense layer to project the difference of candidate embedding pairs into a scalar score:
	\begin{equation}
	\label{eq:3}
	\begin{aligned}
	C_{qr} = \theta_{{di\!f\!\!f}}([f(w_{h^oq}) - f(w_{h^sr})] )+ b_{{di\!f\!\!f}},
	\end{aligned}
	\end{equation}
	where $\theta_{{di\!f\!\!f}} \in{\mathbb{R}}^{d\times1}$ is a weight matrix and $b_{{di\!f\!\!f}}\in{\mathbb{R}}$ a bias term. The score matrix $C = (C_{qr})$ is of dimension $l\times p$. We use row-wise avg-pooling and apply softmax to compute inter mutual attention scores for the original headline:
	\begin{equation}
	\label{eq:5}
	\begin{aligned}
	A^o = \Softmax(\underset{row}{\avg C}) ,
	\end{aligned} 
	\end{equation} 
	where $A_o$ is the learned inter-mutual attention weight vector for original headline.
	We use column-wise avg-pooling and apply softmax to compute inter mutual attention scores for synthetic headline.
	\begin{equation}
	\label{eq:5}
	\begin{aligned}
	A^s = \Softmax(\underset{col}{\avg C}) ,
	\end{aligned} 
	\end{equation} 
	where $A^s$ is the learned inter-mutual attention weight vector for synthetic headline.
	Subsequently, inter-mutual attended representations for original headline $M_{A^o}$ and for synthetic headline $M_{A^s}$ can be computed as:
	\begin{equation}
	\label{eq:6}
	\begin{aligned}
	M_{A^o} &= \sum_{i=1}^{l} f(w_{h^oi}){A^o}_i\\
	M_{A^s} &= \sum_{i=1}^{p} f(W_{h^si}){A^s}_i,
	\end{aligned}
	\end{equation}
	where $f(w_{h^oi})$ and $ f(W_{h^si})$ are the word embeddings of the $i^{th}$ word of the original and synthetic headline, respectively.
	Now, we compute the overall inter-mutual attended representation $M_{A}$ as:
	\begin{equation}
	\label{eq:6}
	\begin{aligned}
	M_{A} = M_{A^o} + M_{A^s}
	\end{aligned}
	\end{equation}

	\subsubsection{Variants of Inter Mutual Attention}
	Although the MuSeM model with difference of word embeddings of candidate word pair works well, we also investigate and experiment with two other versions of MuSeM that use the dot product of candidate word embedding pairs and concatenation of word embedding pairs, similar to \cite{emnlp2018}. We notice that the dot-product and concatenation variants do not perform better than the difference-oriented variant in our experiments. We also combine all three operations, namely difference, dot-product, and concatenation, which performs slightly better than the purely difference oriented model. Attention scores for the dot-product and concatenation oriented models can be computed in a very similar fashion to the difference model, except for equation \eqref{eq:3}, which needs to be replaced by: 
	\begin{equation}
	\label{eq:4}
	\begin{aligned}
	C_{qr} = \theta_{{dot}}([f(w_{h^oq})\cdot f(w_{h^sr})] ) + b_{{dot}}\\
	C_{qr} = \theta_{{con}}([f(w_{h^oq}) \mathbin\Vert f(w_{h^sr})] ) + b_{{con}}.\\
	\end{aligned}
	\end{equation}
	For the sake of brevity, we do not repeat all the other equations. 
	
	\subsubsection{Clubbed Model}
	We also experiment with a combined model in which all three variants namely difference based, dot-product based and concatenation based methods are used in parallel and resulted overall embedding representation is used for computing attention scores. 
	\begin{equation}
	\label{eq:4}
	\begin{aligned}
	F_{qr}^{dot} = [f(w_{h^oq})\cdot f(w_{h^sr})] \\
	F_{qr}^{con}= [f(w_{h^oq}) \mathbin\Vert f(w_{h^sr})]\\
	F_{qr}^{di\!f\!\!f} =[f(w_{h^oq}) - f(w_{h^sr})]\\
	C_{qr} = \theta_{dpc}([F_{qr}^{dot} \mathbin\Vert F_{qr}^{con}\mathbin\Vert F_{qr}^{di\!f\!\!f} ]) + b_{dpc},
	\end{aligned}
	\end{equation}
	where $C_{qr}$ is overall attention score, $\theta_{dpc}$ is the weight matrix and $b_{dpc}$ is the bias term for the clubbed model.
	\begin{figure*}[htbp]
		\centering
		\hspace*{-1mm}\includegraphics[scale=0.21]{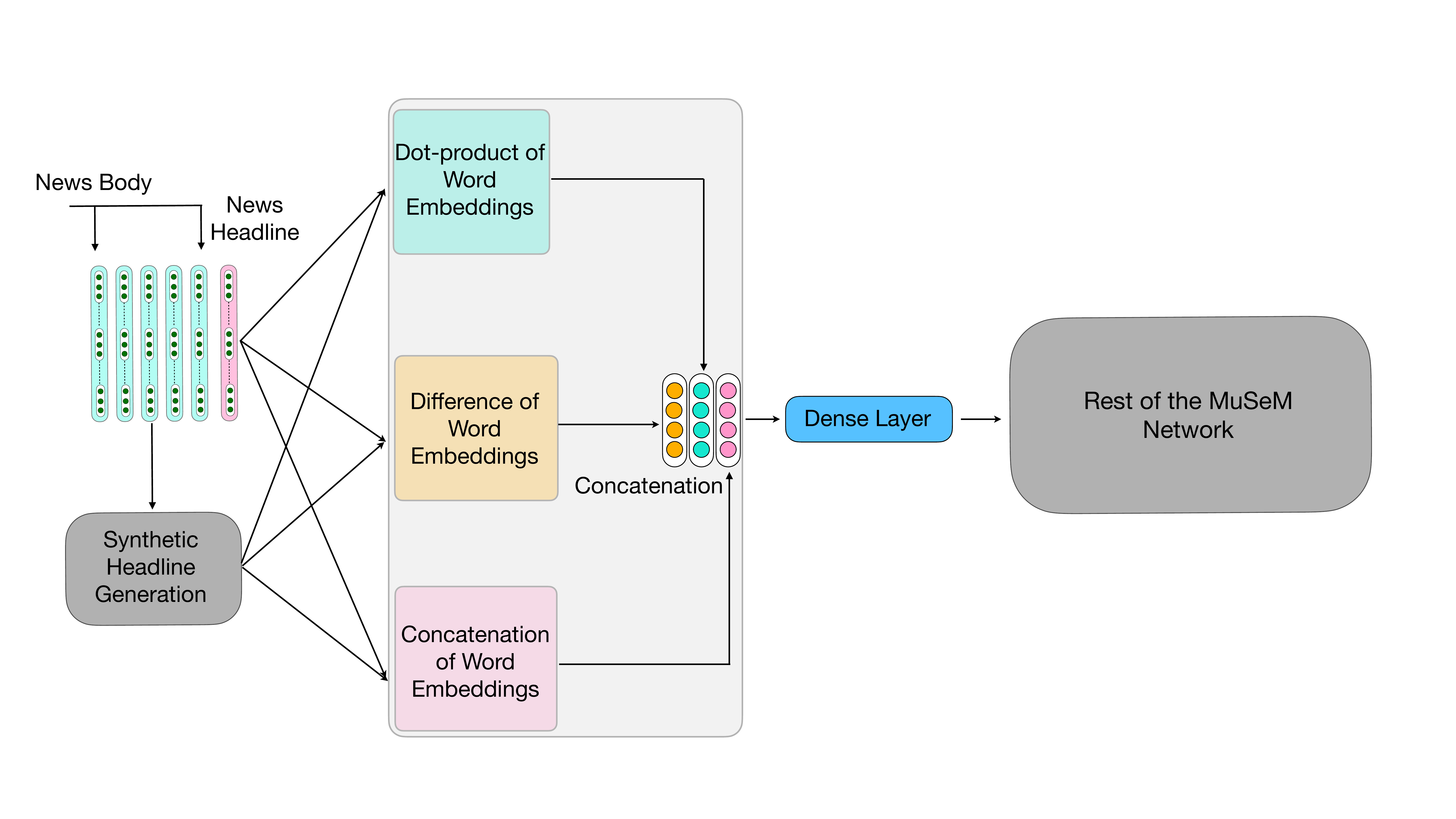}
		\caption{Clubbed Model Architecture }
		
		\label{fig:attention}
		
	\end{figure*}
	\subsection{LSTM based Sequence Encoder}We concatenate the sequence of word embeddings of the words of the original headlines and words of the synthetically generated headline and use long short-term memory unit (LSTM) \cite{Schmidhuber} to encode this overall sequence by using the standard LSTM equations as follows: 
	
	\begin{equation}
	\label{eq:2}
	\begin{aligned}
	f_t = \sigma(W_f.[h_{t-1},x_t]+b_f)\\
	i_t = \sigma(W_i.[h_{t-1},x_t] +b_i)\\
	\tilde{C_t} = \tanh(W_C.[h_{t-1},x_t] +b_C)\\
	C_t = f_t*C_{t-1} + i_t* \tilde{C_t}\\
	o_t = \sigma(W_o.[h_{t-1},x_t] +b_o)\\
	h_t = o_t*\tanh(C_t),
	\end{aligned}
	\end{equation}
	where $h_{t-1}$ is previous hidden state and $x_t$ is the current input. We use the last hidden state of LSTM as the encoded representation of the overall sequence $M_{E}$.

	\subsection{Classification}
	We learn a joint representation $M$ of inter-mutual attended representation $M_{A}$ and encoded representation of the overall sequence $M_{E}$ using a nonlinear transformation layer with ReLU activation. Subsequently, we use a softmax layer to predict the label:
	\begin{equation}
	\label{eq:6}
	\begin{aligned}
	M = \mathrm{ReLU}(W_t ([M_{A},M_{E}]+b_t)\\
	\hat{y} = \Softmax(W_{cl} M + b_{cl}),
	\end{aligned}
	\end{equation}
	where $W_{cl}$, $b_{cl}$, and $\hat{y}$ are the weight matrix, the bias term, and the predicted label, respectively. We use softmax cross-entropy with logits as loss $L$:
	\begin{equation}
	\label{}
	\begin{aligned}
	L = -{\sum}^m_{i=1}\log\frac{e^{w^T_{y_i}x_i+b_{y_i}} }
	{\sum\limits_{j=1}^{n}e^{w^T_j x_i+b_j}}.
	\end{aligned}
	\end{equation}

	\section{Experimental Setup}
	We implement the proposed models within the TensorFlow framework. For the evaluation and comparison, we use Macro F1 and AUC scores as evaluation metrics. All the parameters are tuned using a grid search. As a result, we keep learning rate as 0.001, batch size as 100, hidden states of LSTM as 100, and pretrained GloVe\cite{Pennington14glove:global} embeddings of 300 dimensions. We use softmax cross-entropy with logits as the loss function, maximum sentence length as 50, dropout rate as 0.2, and the number of epochs as 10.  For the synthetic headline generation, we use the default parameter values used in the papers mentioned above. 
	
	\subsection{Dataset Statistics}
	
	We use two publicly available datasets,  NELA17\footnote{\url{https://github.com/BenjaminDHorne/NELA2017-Dataset-v1}} and Click-bait Challenge\footnote{\url{http://www.clickbait-challenge.org/}} for the evaluation and comparison of the proposed model with the baseline methods. The NELA17 dataset is provided by \cite{Yoon}. Although they do not evaluate their model with it, they provide a script\footnote{\url{https://github.com/sugoiii/detecting-incongruity-dataset-gen}} to generate the dataset from an original news collection dataset. It contains 91042 news items in total, of which 45521 news items are congruent, and 45521 news items are non-congruent. The Click-bait Challenge dataset is a collection of social media posts, which are annotated as click-bait or non-click-bait using a crowd-sourcing platform via majority voting. It contains 21033 social media posts in total, of which 16150 posts are congruent, and 4883 posts are non-congruent. 
	\begin{table}[htbp]
		\centering
		\caption{Dataset Statistics}
		
		
		\begin{minipage}{0.5\textwidth}
			\centering
			\scalebox{1.2}{
				\begin{tabular}{||p{2cm}|p{3cm}||}

					\hline
					\bf Statistics    &\bf NELA17 \\\hline 
					Total &91042\\\hline 
					Non-congruent &45521\\\hline
					Congruent&45521\\\hline
				\end{tabular}
			}
		\end{minipage}
	\end{table}
	\begin{table}[htbp]
		\centering
		\begin{minipage}{0.5\textwidth}
			\centering
			\scalebox{1.2}{
				\begin{tabular}{||p{2cm}|p{3cm}||}

					\hline
					\bf Statistics    &\bf Click-bait Challenge \\\hline 
					Total &21033\\\hline 
					Non-congruent &4883\\\hline
					Congruent& 16150\\\hline
				\end{tabular}
			}
		\end{minipage}
		
	\end{table}       
	
	The NELA17 dataset is balanced but the Click-bait Challenge dataset has class imbalance problem. We use the class\_weight\footnote{\url{https://www.tensorflow.org/tutorials/structured\_data/imbalanced\_data}.} parameter to adjust the imbalance in the dataset.

	\section{Evaluation and Discussion}
	In this section, we compare the performance of the proposed models with state of the art methods and baseline models on two publicly available datasets. Subsequently, we also provide some explanations on the differences in the performance of the analyzed models. We also discuss a potentially useful trick from \cite{Mishra_2020_CVPR_Workshops}, which can improve the incongruence detection accuracy by capturing the newly uncovered cues. 
	
	We compare the proposed model with these baselines:
	\begin{itemize}
		\item \textbf{SVM:}\cite{cortes1995support} In this method, we use handcrafted linguistic and other statistical features to learn a classifier with support vector machine technique. 
		\item \textbf{LSTM:}\cite{Schmidhuber} In this method, we use long short term memory unit to encode both headline and body pair and apply softmax for the classification. 
		\item \textbf{Hi-LSTM:}\cite{tang-etal-2015-document} This technique uses an LSTM based hierarchical encoder, which first encodes words and then uses another LSTM encoder to form the sentence representations.
		\item \textbf{Yoon:}\cite{Yoon} This is a state of the art, hierarchical dual encoder based model which uses headline guided attention to learn the contextual representation.
	\end{itemize}
	We experiment with four variants of the proposed MuSeM model.
	\begin{itemize}
		\item \textbf{MuSeM\_diff\_SeqGAN:} This is the MuSeM model with the difference between word embeddings, and synthetic headlines are generated using SeqGAN method.
		\item \textbf{MuSeM\_dpc\_SeqGAN:} This is the MuSeM model which uses a combination of all three operators, namely difference, concatenation, and dot-product between word embeddings and synthetic headlines are generated using SeqGAN method.
		\item \textbf{MuSeM\_diff\_SHG:}This is the MuSeM model with the difference between word embeddings, and synthetic headlines are generated using SHG method. 
		\item \textbf{MuSeM\_dpc\_SHG:} This is the MuSeM model which uses a combination of all three operators, namely difference, concatenation, and dot-product between word embeddings and synthetic headlines are generated using SHG method.
	\end{itemize}
	\begin{table}[htbp]
		\centering
		\caption{Comparison of the proposed models with various state of the art baseline models for the NELA17 Dataset.}
		\begin{minipage}{0.5\textwidth}
			\centering
			\centering
			\scalebox{1.2}{
				\begin{tabular}{||p{3cm}|p{1cm}|p{1cm}||}
					
					\hline
					\multicolumn{3}{||c||}{NELA17 Dataset}\\
					\hline	\hline
					\bf Model & \bf   Macro F1 & \bf AUC. \\\hline 
					
					SVM &0.622 &0.637\\\hline 
					LSTM&0.642&0.663\\\hline 
					HiLSTM&0.651&0.672\\\hline 
					Yoon &0.685& 0.697\\\hline 
					MuSeM\_diff\_SeqGAN & 0.713 &0.720\\\hline 
					MuSeM\_dpc\_SeqGAN & 0.719 &0.727\\\hline 
					MuSeM\_diff\_SHG &0.740 &0.753\\\hline 
					MuSeM\_dpc\_SHG &\textbf{0.752} &\textbf{0.769}\\\hline
					
				\end{tabular}
			}%
		\end{minipage}
	\end{table}		
	\begin{table}[htbp]
		\centering
		\caption{Comparison of the proposed models with various state of the art baseline models for the Click-bait challenge 2017 Dataset.}
		\begin{minipage}{0.5\textwidth}
			\centering
			\scalebox{1.2}{
				\begin{tabular}{||p{3cm}|p{1cm}|p{1cm}||}
					
					\hline
					\multicolumn{3}{||c||}{Click-bait challenge 2017 Dataset}\\
					\hline	\hline
					\bf Model & \bf  Macro F1 & \bf AUC. \\\hline 
					
					SVM &0.618 &0.629\\\hline 
					LSTM&0.630&0.641\\\hline 
					HiLSTM&0.642&0.656\\\hline 
					Yoon &0.660&0.678 \\\hline 
					MuSeM\_diff\_SeqGAN &0.677 &0.683\\\hline 
					MuSeM\_dpc\_SeqGAN & 0.690&0.698\\\hline
					MuSeM\_diff\_SHG &0.729 &0.734\\\hline 
					MuSeM\_dpc\_SHG &\textbf{0.735} &\textbf{0.747}\\\hline 
					
				\end{tabular}
			}
		\end{minipage}
	\end{table}		
	
	\subsection{Results for NELA17 Dataset}
	In the case of the NELA17 dataset, the SVM model with linguistic and statistical features achieves $0.622$  and $0.637$ in terms of Macro F1 and AUC, respectively. The LSTM model outperforms SVM with Macro F1 as $0.642$ and AUC as $0.663$. This gain can be attributed to the suitability of LSTM to learn the contextual representation of text sequences. The Hierarchical LSTM (Hi-LSTM) performs slightly better than simple LSTM with Macro F1 as $0.651$ and AUC as $0.672$. The probable reason for this improvement is the better representation learned by H-LSTM, in the form of the documents' hierarchical structure. Since the Yoon model uses a dual hierarchical encoder, which encodes words and paragraphs of the new body text separately using an attention mechanism guided by news headline, it outperforms the Hi-LSTM with significant gains. The Yoon\footnote{\url{https://github.com/david-yoon/detecting-incongruity}} model works well for long texts as it selects important paragraphs from the long body text, which reduces the effective size of the document. On the other hand, SVM, LSTM, and Hi-LSTM models do not scale well for long text sequences.
	
	We compare the baselines and state-of-the-art models with four variants of the proposed model MuSeM. The MuSeM\_diff\_SeqGAN model performs better than the Yoon model with $0.713$ and $0.720$ in terms of Macro F1 and AUC. This performance improvement can be credited to low dimension representation of the news body content in the form of synthetic headline and inter-mutual attention based semantic matching. Although the headline guided attention to select relevant paragraphs in the Yoon model reduces the effective document length, the resultant document representation is still not of very low dimension. In contrast, the MuSeM model uses a very low dimensional representation of news body content in the form of a synthetic headline, which is more effective in semantic matching. 
	
	MuSeM captures both compositional aspects and latent cues related to similarity and dissimilarity relationships among the words using LSTM based sequence encoder and inter-mutual attention based semantic matching. The MuSeM\_dpc\_SeqGAN model performs slightly better than  MuSeM\_diff\_SeqGAN due to additional patterns pertaining to similarity relationship among words, captured by the combination of all three variants, i.e., difference based, dot-product based, and concatenation based methods. The MuSeM\_diff\_SHG model, which uses a style transfer based headline generation method, outperforms both variants of the MuSeM models with SeqGAN by a large margin. This performance improvement can be attributed to a better headline generation by stylized headline generation (SHG) method. The MuSeM\_dpc\_SHG model outperforms all the other models by achieving $0.752$  and $0.769$ in Macro F1 and AUC, respectively.

	\subsection{Results for Click-bait Challenge Dataset}
	
	In the case of the Click-bait Challenge dataset, we observe very similar trends as with the NELA17 dataset. The deep learning based methods perform significantly better than the handcrafted feature based SVM model. With  $0.660$ and $0.678$ in Macro F1 and AUC, the Yoon model outperforms both the LSTM and Hi-LSTM models. All variants of the MuSeM model perform better than all the baseline methods, whereas MuSeM\_dpc\_SHG model achieves $0.735$  and $0.747$ in terms of Macro F1 and AUC, beating all the other variants.

	\subsection{Higher Order Inter-mutual Attention}
	Authors in \cite{Mishra_2020_CVPR_Workshops} propose a variant of mutual attention which can be used to model higher-order relationship among the candidate words. Essentially, the proposed inter-mutual attention only models the relationship between two words at a time individually, and the presence of other words in the sentences are not taken into account. If we can also reckon the presence of other words during the computation of the inter-mutual attention, we can capture the additional contextual cues related to similarity or dissimilarity relationship. The higher-order mutual attention trick allows us to model such relationships. 
	
	There are certain challenges to be taken care of before the application of higher order mutual attention to incongruence detection problem. Firstly, the original higher order mutual attention is designed for a single sequence as it uses intra-mutual attention within a sequence, while headline incongruence detection involves two sentences or sequences. Secondly, higher order mutual attention is achieved by applying a square or cube of the attention score matrix, which is possible only if the original score matrix is a square matrix. In the headline incongruence detection task, the score matrix may or may not be a square matrix because original and synthetically generated headlines may have different lengths.

	\section{Conclusion}
	This paper proposes inter-mutual attention based semantic matching (MuSeM) to detect incongruence in news headlines. We also investigate a different variant of MuSem, which combines three operations on word embedding pairs to compute inter-mutual attention scores. The proposed models outperform all the baselines in experiments with two publicly available datasets. We notice that the performance of inter-mutual attention based semantic matching greatly depends on the accuracy of synthetic headline generation step. 
	
	In future research, we plan to use the higher order word to word attention trick used in \cite{Mishra_2020_CVPR_Workshops} to model and capture the indirect relationships between the word pairs. We plan to investigate and devise an end to end version of MuSeM model, where synthetic headline generation and semantic matching steps are seamlessly integrated. We also plan to conduct an analysis of the efficacy of the attention mechanism by visualizing the attention weights.

	\bibliographystyle{plain}

\end{document}